%% file: main.tex
\documentclass[11pt]{article}
\usepackage{amsmath,amssymb}
\usepackage{primeintellect/primeintellect_preprint}
\usepackage{booktabs,array,multirow,tabularx}
\usepackage{subcaption}
\usepackage{enumitem}
\usepackage{xspace}
\usepackage{listings}
\usepackage{float}

\makeatletter
\def\fps@figure{!t}
\makeatother

\newcommand{\rlmcall}{\texttt{rlm}\xspace}
\newcommand{\code}[1]{\texttt{#1}}
\definecolor{codebg}{HTML}{F4F5F6}
\title{Prime Agent: A Self-Improving RLM Harness}
\author{
Seth Karten\textsuperscript{\(\spadesuit\,\heartsuit\,\diamondsuit\)} \quad
Alex L. Zhang\textsuperscript{\(\heartsuit\,\clubsuit\,\diamondsuit\)} \quad
Kevin Thomas\textsuperscript{\(\heartsuit\)} \quad
Sebastian M\"uller\textsuperscript{\(\heartsuit\)} \quad \\[0.2em]
Elie Bakouch\textsuperscript{\(\heartsuit\)} \quad
Daniel Auras\textsuperscript{\(\heartsuit\)} \quad
Mika Senghaas\textsuperscript{\(\heartsuit\)} \quad
Fares Obeid\textsuperscript{\(\heartsuit\)} \quad\\[0.2em]
Konstantin Dunas\textsuperscript{\(\heartsuit\)} \quad
Johannes Hagemann\textsuperscript{\(\heartsuit\)} \quad
Sami Jaghouar\textsuperscript{\(\heartsuit\)}\\[0.45em]
{\small \textsuperscript{\(\spadesuit\)}\,Princeton University \qquad \textsuperscript{\(\heartsuit\)}\,Prime Intellect \qquad \textsuperscript{\(\clubsuit\)}\,MIT}\\
{\small \textsuperscript{\(\diamondsuit\)}\,Correspondence: \texttt{seth@primeintellect.ai}, \texttt{altzhang@mit.edu}}
}
\date{First published: August 5, 2026 \quad\textbullet\quad Current version: August 24, 2026}
\runningtitle{Prime Agent: A Self-Improving RLM Harness}
\venuetag{Technical Report}

\begin{document}
\maketitle
\input{sections/00_abstract}
\input{sections/01_introduction}
\input{sections/02_methodology}
\input{sections/03_evaluation}
\input{sections/04_related_work}
\input{sections/05_conclusion}

\section*{Acknowledgements}
We thank Florian Brand, SinatraS, and Patience Cave for helping complete Prime Agent runs in additional environments.
SinatraS created PMPP Hard, and Patience Cave created MazeBench.

\bibliography{references}
\bibliographystyle{plainnat}
\appendix
\input{sections/A_appendix}
\end{document}

%% file: sections/00_abstract.tex
\begin{abstract}
Language models are sequential processors, but long-horizon agency requires external information and computation beyond model weights and active context.
Prime Agent is an open-source harness for long-horizon evaluation and coding-agent workflows.
A persistent IPython REPL follows the Recursive Language Model abstraction for programmatic context processing and test-time compute, while Continual Harness preserves histories, memories, skills, prompts, and subagent specifications across trajectories.
Recursive subagents coordinate through direct agent-to-agent communication, and the Agents View lets humans inspect and manage daemon-backed sessions.
Prime Agent standardizes execution, recovery, verification, and resource accounting while leaving strategy construction to the model.
This low-friction, expressive membrane prevents harness failures from becoming model failures and pushes measurement toward the model's true maximal underlying capability.
Prime Agent raises ARC-AGI-3 RHAE Best@1 from 30\% to 95.5\% and matches or exceeds native and popular harnesses across long-context coding, GPU-kernel generation, emulator construction, and autonomous nanoGPT speedruns.
On Factorio, we find refinement allows for continuous technology progression and dedicated subagents enable parallelized work.
Code is available at \url{https://github.com/PrimeIntellect-ai/prime-agent}.
\end{abstract}

%% file: sections/01_introduction.tex
\section{Introduction}
A strong language model on its own does not have the full capabilities of a computer.
An LLM is a bounded sequential processor whose next decision can use only state information exposed in its weights and active context.
A harness supplies the missing computational substrate that allows for external actions via tool-calls.
These external actions also include information management beyond prior knowledge stored in the model weights. Context management was first enabled by agentic compaction, the process by which a model selectively analyzes its own context to reduce tokens while keeping essential information.
However, the full information state has grown beyond the weights and token context.
Imagined as a state information cache (\cref{fig:llm-hierarchy}), model weights are L0, active context is L1, a persistent REPL and recursive subagents form L2, and disk-backed history, memories, and skills form L3.
This makes the system more \emph{von Neumann-like}: the model can read, transform, and write addressable state outside the instruction currently being generated~\citep{turing1936computable,vonneumann1945edvac}.

This perspective makes \emph{expressivity} the key property of a harness.
Rather than encode one workflow, an expressive harness exposes primitives from which the model constructs programs, subagents, and feedback loops at inference time.
Recursive Language Models (RLMs) make context and recursive invocation programmable~\citep{zhang2025recursivelanguagemodels}; Continual Harness makes prompts, subagents, skills, and memories revisable from the trajectory history~\citep{karten2026continualharness}.
Lastly, we enable large-scale coordination and orchestration of multi-agent swarms through direct agent-to-agent communication.
These components let a fixed model use information management and test-time compute to expand its reachable strategy set.

Prime Agent is designed first as a standardized harness for \emph{long-horizon evaluation}.
The most popular metric of which is score at a fixed expenditure~\citep{metr2026}, such as cost/tokens and time, or, especially for long horizon, score at practical plateau~\citep{metr2026}, which allows us to analyze the shape of performance over time.
The harness is the membrane through which the model observes and acts on the world.
A model should fail an evaluation because the task exceeds its capability, not because the harness dropped state, restricted useful actions, miscounted resources, or terminated prematurely.
Prime Agent therefore combines standardized, reliable execution with a low-friction and expressive interface for programmatic tools, information management, and swarm management.
This lets the model fully use its test-time compute, pushing measured performance toward its true maximal underlying capability rather than the limitations of its harness.

Prime Agent jointly manages information and computation.
Information management moves state across L1--L3 through programmatic context processing, compaction, persistent histories, and revisable memories.
Computation management allocates test-time compute to programs, tool calls, reusable skills, and parallel recursive subagents~\citep{snell2024scalingtesttime,zhang2025recursivelanguagemodels}.
Direct agent-to-agent communication connects the two, routing information across distributed computation so the swarm can coordinate dynamically rather than follow a fixed graph.
These communication links also let humans inspect, message, attach to, and intervene in subagent sessions without following every exchange~\citep{karten2023interpretablecommunication,li2023theoryofmind}.
Together, these interactions produce retained trajectories that improve future computation and can train later model generations~\citep{zelikman2022star,xu2024agenttrek,pan2024swegym}.

We present Prime Agent, an open-source harness for long-horizon model evaluation and coding-agent workflows.
Prime Agent integrates information and computation management across active context, persistent programmatic execution, recursive subagents, and retained histories, memories, and reusable skills, connected through direct agent-to-agent and human-agent communication.
Its Agents View provides a visual interface for inspecting, attaching to, and managing persistent daemon-backed agent sessions.
We standardize evaluation infrastructure while preserving the model's freedom to construct its own strategy.
Prime Agent improves ARC-AGI-3 performance from 30\% to 95\%, matches or exceeds Pi, Claude Code, and Codex (and outperforms other harnesses, Hermes Agent, OpenCode, and Kimi-Code, on other benchmarks) across long-context coding, GPU-kernel generation, and emulator construction, sustains an 85.5-hour nanoGPT run with 19 validated records, and supports four-character Factorio control and long-horizon MazeBench exploration.

%% file: sections/02_methodology.tex
\section{Prime Agent Architecture}
\label{sec:architecture}
\begin{figure}[!t]
\centering
\includegraphics[width=\linewidth]{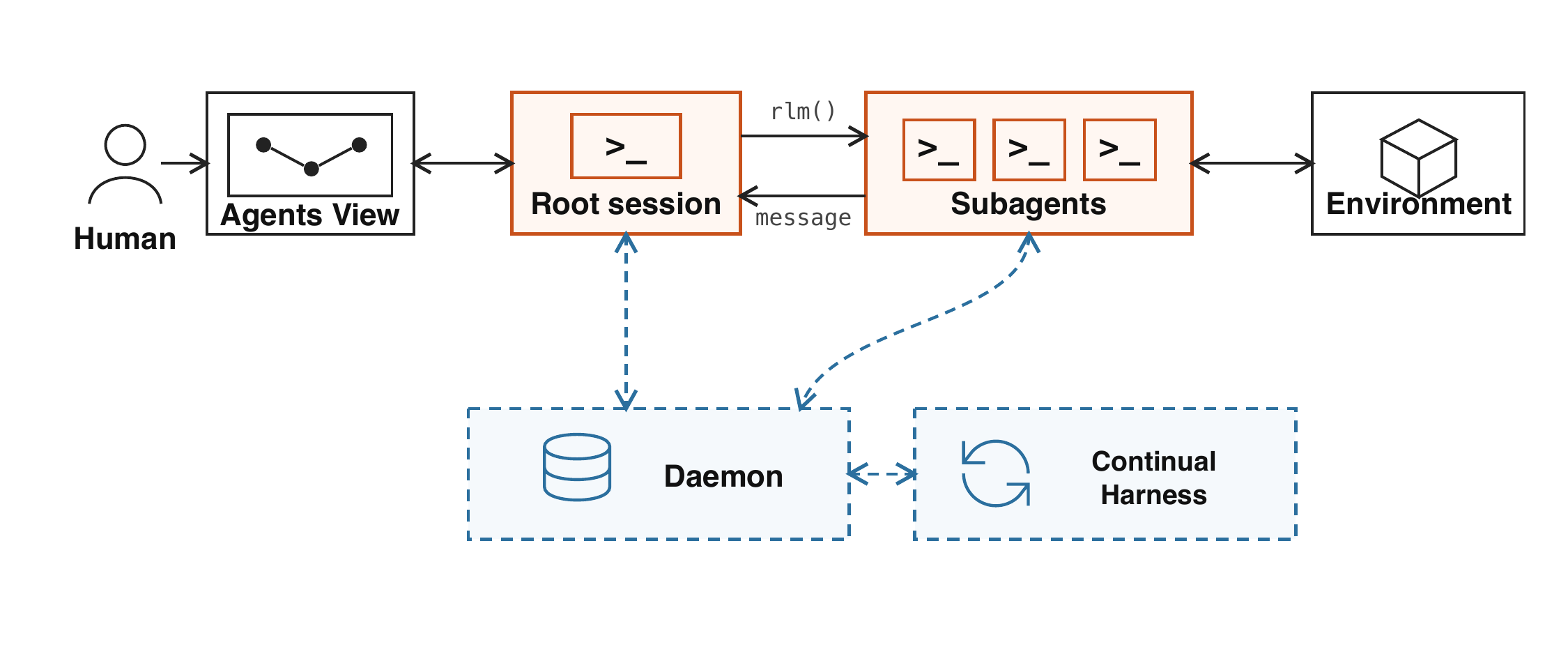}
\caption{Prime Agent connects persistent root and subagent sessions to a daemon, Continual Harness, the Agents View, and the environment. Solid arrows carry execution and messages; dashed arrows carry persistent state.}
\label{fig:architecture}
\end{figure}

We next describe the Prime Agent architecture in detail.
In particular, we discuss (1) how the system manages information and computation (\S\ref{sec:architecture-overview}), (2) how state is organized across model weights, active context, the REPL, and disk-backed storage (\S\ref{sec:information-hierarchy}), (3) how persistent REPLs and RLM calls support programmatic computation (\S\ref{sec:programmatic-computation}), (4) how recursive sessions communicate with agents and human operators (\S\ref{sec:recursive-orchestration}), (5) how Continual Harness turns trajectory evidence into reusable state (\S\ref{sec:continual-harness}), and (6) how long-horizon controls define continuation, termination, and evaluation accounting (\S\ref{sec:long-horizon-execution}).
\Cref{fig:architecture} provides an overview.

\subsection{Architecture overview}
\label{sec:architecture-overview}
Prime Agent separates information management from computation management.
Information management determines what state enters a model invocation and what survives compaction or restart.

Computation management maps model-selected actions to code, tools, and recursive subagent sessions.
Direct agent-to-agent communication connects related sessions, and direct human-agent interaction exposes individual nodes for inspection and intervention.

Models manipulate intermediate values with code.
Sessions retain history across compaction, detachment, and restart.
Subagents inherit the root's execution and communication primitives.
The runtime records model calls, tool use, messages, harness changes, and resource use.
The model controls decomposition, computation allocation, communication, and stopping.

\subsection{Information hierarchy and persistent state}
\label{sec:information-hierarchy}
\begin{figure}[H]
\centering
\includegraphics[width=\linewidth]{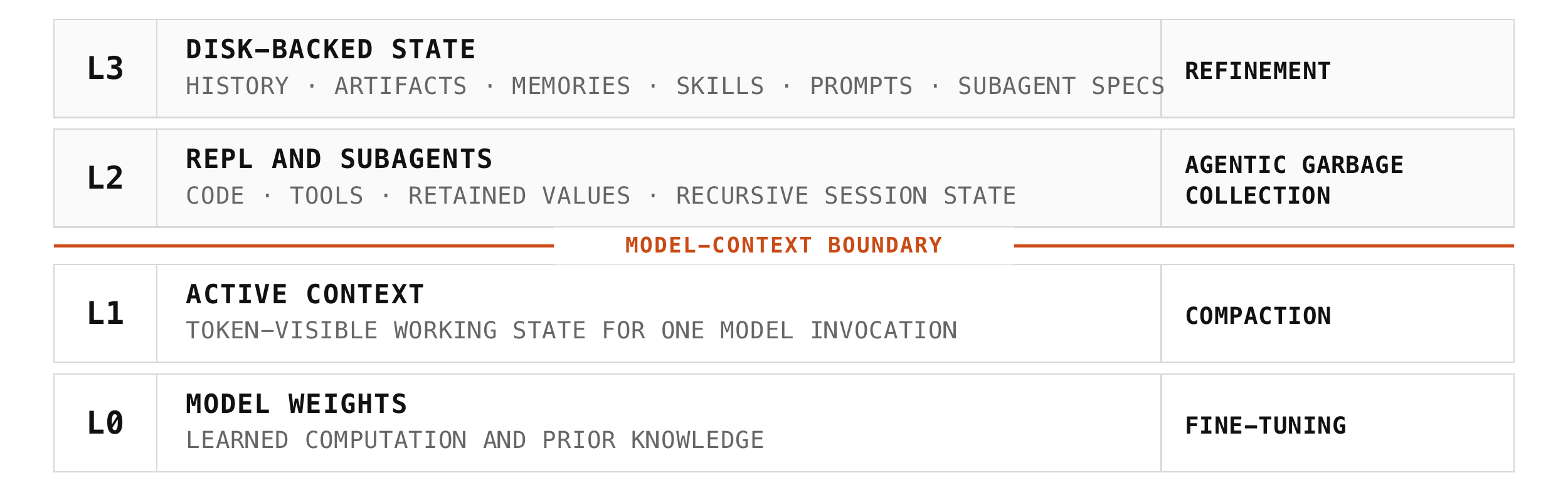}
\caption{\textbf{Prime Agent state hierarchy.} The boundary between L1 and L2 separates token-visible model state from explicitly managed computation and retained state.}
\label{fig:llm-hierarchy}
\end{figure}

A model invocation is parameterized by fixed weights and conditions on its active token context.
Prime Agent adds state outside that context. External state affects generation when the runtime injects it or an operation serializes a result into the context.
\Cref{fig:llm-hierarchy} organizes this state by visibility, access mechanism, and persistence.

Each level changes through a different mechanism.
Fine-tuning updates L0, compaction rewrites L1, and refinement versions selected L3 entries.
We call the L2 mechanism \emph{agentic garbage collection}.
The model creates, retains, summarizes, or deletes REPL values and subagent sessions as the task changes.

Explicit operations move information between levels.
Python values and tool outputs in L2 enter generation when serialized into L1.
Compaction replaces a conversational prefix with a summary and retains the original events in L3 for REPL retrieval.
The runtime assembles selected Continual Harness entries into later supplemental prompts; other L3 artifacts enter context on retrieval.
L0 stays fixed.

The retained runtime state includes an append-only event history, selected kernel snapshots, the rooted session tree, context and compaction records, persistent message queues, and versioned Continual Harness state.
Branching or forking creates a new logical continuation without deleting the prior event sequence.
Recovery reconstructs the session under the same identity.
Non-serializable Python objects and external processes are recreated from saved artifacts or external services.

\subsection{Programmatic computation with RLMs}
\label{sec:programmatic-computation}
Each session owns a persistent IPython Read-Eval-Print Loop (REPL).
At test time, compute comprises model inference, Python execution, and tool calls; evaluations report tokens, time, and cost separately.
Installed tools are imported as Python modules for parsing, filtering, aggregation, and verification with ordinary code.
Intermediate values persist across turns and remain outside active context until selected.
This avoids repeatedly serializing large logs, task specifications, and structured evaluator output into the context.

Prime Agent implements the RLM abstraction with the asynchronous \rlmcall primitive~\citep{zhang2025recursivelanguagemodels}.
Calling \rlmcall creates and schedules a subagent session, then returns a stable handle before the subagent completes.
The subagent receives its own model context, IPython kernel, history, and workspace metadata.
The parent continues local computation while subagents run.
Results arrive later through direct agent-to-agent communication, and retained handles support follow-up after compaction or restart.
The model chooses between local code, tools, sequential delegation, and parallel subagents.
Prime Agent defines their execution semantics instead of a fixed workflow graph.
A complete orchestration example appears in \cref{app:orchestration}.

\subsection{Recursive orchestration and interaction}
\label{sec:recursive-orchestration}
The daemon owns live sessions independently of the client that created them.
Root and subagent sessions use the same lifecycle.
Sessions are running during a turn or tool operation, idle when loaded without an active turn, and inactive when unloaded but recoverable from persistent state.
Client detachment leaves the session running.
Stable session and parent identifiers preserve the recursive topology across these transitions.

\begin{figure}[!t]
\centering
\includegraphics[width=\linewidth]{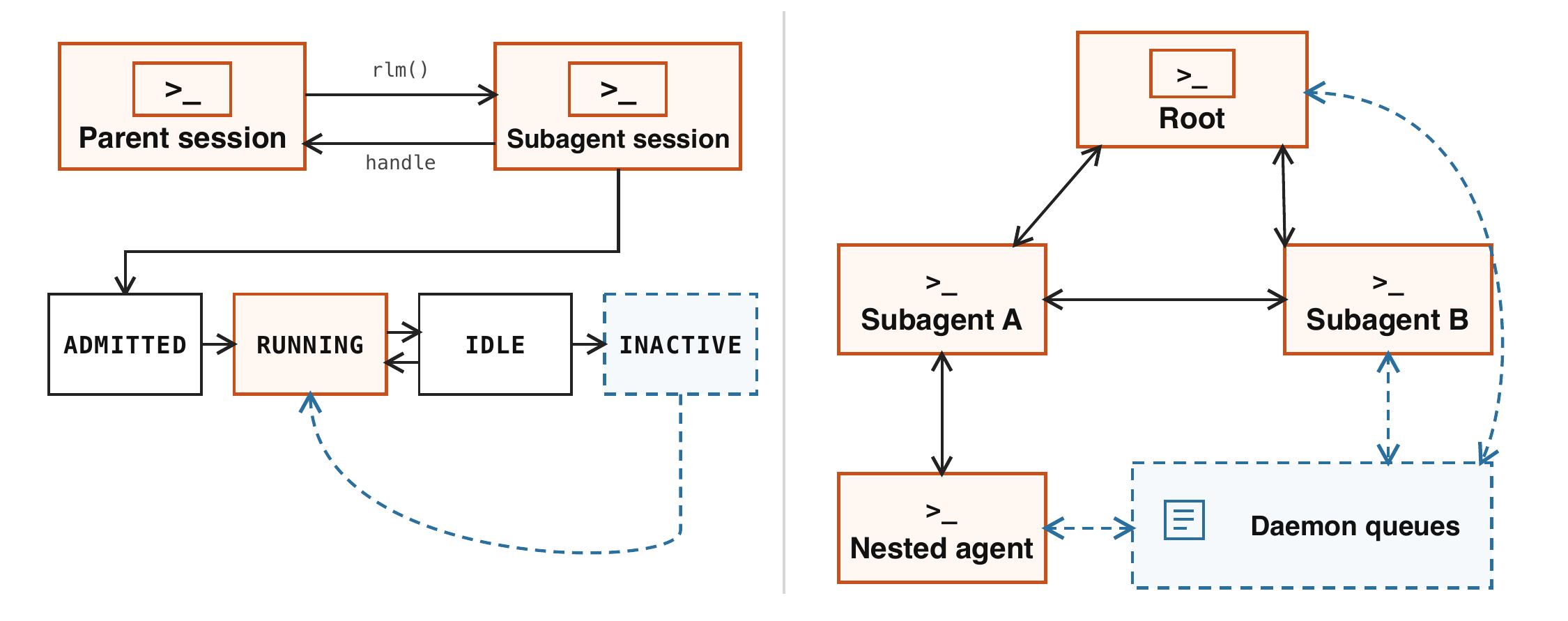}
\caption{Multi-agent orchestration lifecycle and direct agent-to-agent communication.}
\label{fig:orchestration-lifecycle}
\end{figure}

Direct agent-to-agent communication uses asynchronous, daemon-mediated queues.
An agent addresses its parent, children, and siblings, and queued messages remain available when a recipient becomes active again.
Filesystem, network, and credential access follow the permissions of the runtime environment.
\Cref{fig:orchestration-lifecycle} summarizes the shared lifecycle and communication topology.

The Agents View exposes the persistent tree for direct human-agent interaction.
The interface lets a user inspect history, attach to a session, provide new input, or detach without interrupting execution.
The \code{agent-observe} interface provides bounded read-only status and recent-message previews; \code{agent-message} targets a named related session.
This allows for full interaction via the orchestrator.

\subsection{Continual Harness}
\label{sec:continual-harness}
Continual Harness exposes supplemental state for trajectory-time reads and writes~\citep{karten2026continualharness}.
Prompt notes store behavioral instructions, memories store facts, skills package executable procedures, and subagent specifications store reusable roles or divisions of labor.
Typed state separates rules, facts, programs, and coordination patterns.
Entries support create, read, update, and delete operations; local entries belong to one session, and explicitly requested global entries remain available to later sessions.

Refinement converts trajectory evidence into versioned state updates.
Agents request edits directly, or \code{/refine} runs a background model call over relevant events.
The runtime applies each edit at a turn boundary, records its trigger and intended effect, and assembles supplemental state for the next invocation.
Versions preserve provenance and enable rollback.
Refinement supplements the immutable base prompt without rewriting foundational policy.

\emph{Self-improvement} converts execution evidence into persistent harness state that changes later behavior while model weights remain fixed.
Useful computations become skills, repeated coordination patterns become subagent specifications, and corrected assumptions become memories or prompt notes.
The resulting trajectory record also provides training data for later model generations.

\subsection{Long-horizon execution and evaluation semantics}
\label{sec:long-horizon-execution}
\begin{figure}[!t]
\centering
\includegraphics[width=\linewidth]{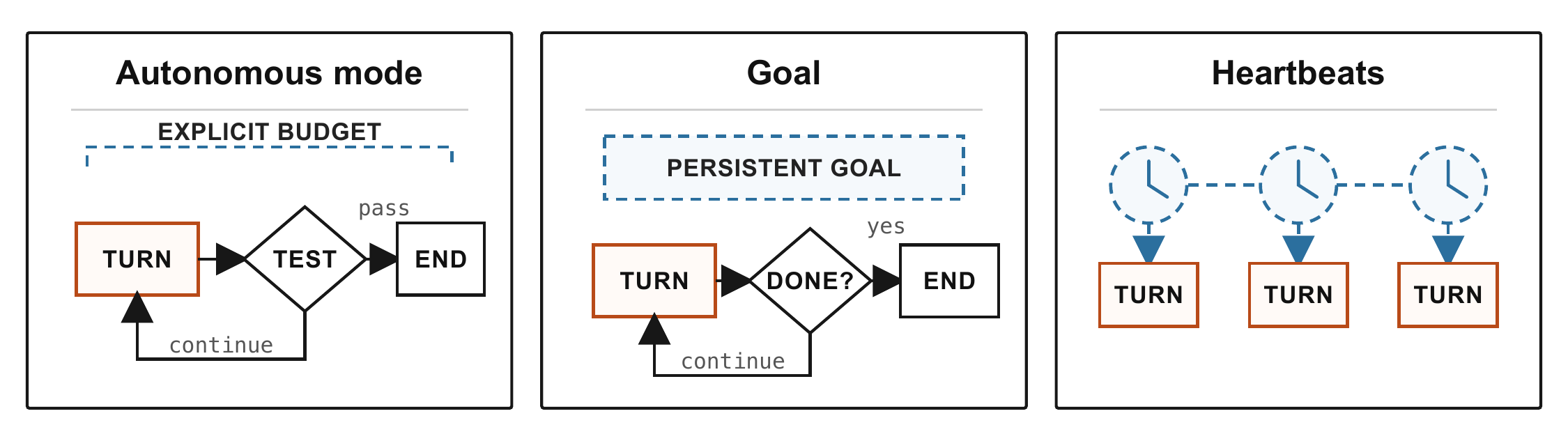}
\caption{\textbf{Long-horizon control mechanisms.} Autonomous mode continues under an explicit budget until an end-condition test passes. Goals retain an objective across continuations until agentic completion. Heartbeats initiate cron or timed turns.}
\label{fig:long-horizon-controls}
\end{figure}

Prime Agent exposes three long-horizon control mechanisms (\cref{fig:long-horizon-controls}).

Autonomous mode continues model turns within an explicit budget and evaluates a task-specified end-condition test after each turn.
A failed test returns bounded output for another attempt; turn, token, and wall-clock limits stop execution.
A goal retains an objective across continuations and ends through agentic completion, when the agent marks the goal complete.
Heartbeats initiate turns on cron or timed schedules.

Evaluation configurations bind task and tool interfaces to model and provider settings, compaction and refinement policies, retry policy, completion gates, and resource limits.
Accounting aggregates the root and descendant sessions, so delegation remains visible in test-time cost.
Event history links model and tool calls, messages, interventions, retries, verifier outcomes, and harness edits to that configuration.
Standardized persistence, recovery, termination, and accounting separate harness failures from model failures while preserving model control over decomposition.

%% file: sections/03_evaluation.tex
\section{Evaluation}
\label{sec:evaluation}
The evaluation addresses three research questions derived from Prime Agent's design.
\textbf{RQ1: Test-time scaling.}
Can a standardized, expressive execution interface let frontier models convert additional output tokens and API cost into verified task progress?
We evaluate this question on ARC-AGI-3.
\textbf{RQ2: Information management.}
Can models use persistent REPL state to search, transform, and aggregate information across long contexts?
We compare Prime Agent with native and alternative harnesses on long-context reasoning and coding tasks.
\textbf{RQ3: Persistent recursive execution.}
Can the same runtime sustain multi-day experimentation, iterative systems construction, recursive control, and online refinement?
We study nanoGPT, PMPP-Hard, EmulatorBench, Factorio, and MazeBench through end-to-end outcomes and trajectory analysis.
The trajectory analyses show how agents allocate subagents, retain information, and recover from disruption.

\subsection{Interactive reasoning at test-time scale}
ARC-AGI-3 is the clearest test of Prime Agent to support strong, consistent long-horizon evaluation~\citep{arcagi3}.
Each game requires the model to learn the rules of the game, creating an ad-hoc world model under an action limit.
Prime Agent supplies only the environment interface and an autonomous prompt adapted from PRO-LONG~\citep{prolong2026}; the model constructs the strategy. We note that Claude Code and Codex runs perform worse than Anthropic and Open AI self-reported performance on ARC-AGI-3 (public set), so we defer to their results over our own runs with matched prompt and settings.

\begin{figure}[!t]
\centering
\includegraphics[width=\linewidth]{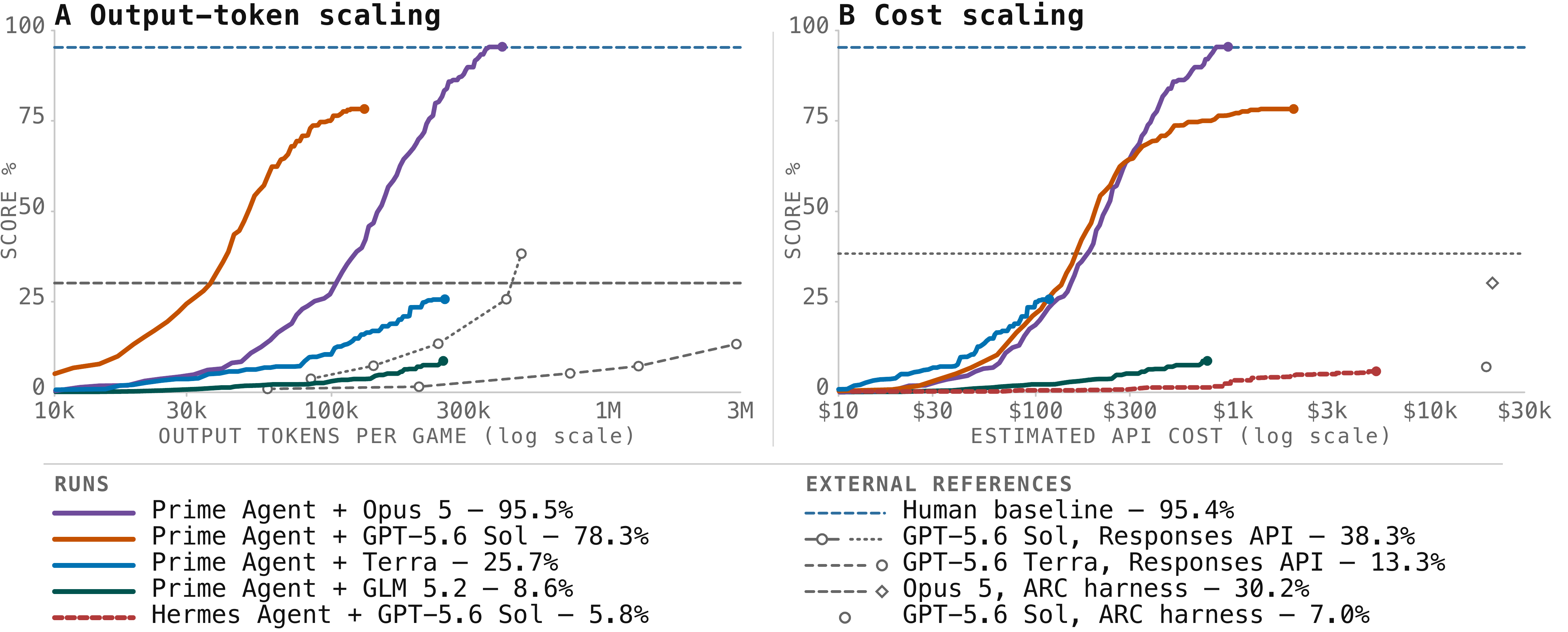}
\caption{\textbf{ARC-AGI-3 test-time scaling.} RHAE score versus output tokens per game (left) and estimated API cost (right).}
\label{fig:arc}
\end{figure}

Across the observed configurations, additional output tokens and cost are converted into progress at sharply different rates (\cref{fig:arc}).
The stronger configurations continue to improve across a long interaction horizon, while others plateau early.
This pattern is consistent with a model-controlled interface that permits model-dependent test-time scaling instead of imposing one fixed workflow.
The reference lines and points place these curves beside official ARC results.
They are external values because our native-harness reruns fell below the published scores, so they situate the result rather than isolate a causal harness effect.

\subsection{Long-context information management}
The long-context suite tests whether a model can actively manage information that does not fit naturally into one prompt.
Prime Agent stores the initial context in a readable file, allowing the model to search, transform, summarize, and revisit it from the persistent REPL.
This changes long-context reasoning from passive attention over a fixed sequence into a programmatic information-management problem.
The suite covers aggregation, latent retrieval, instruction following, reasoning, and long-form coding~\citep{oolong2025,obliq2026,longbenchpro2026,longbenchv2,longcot2026}.

\begin{table}[H]
\centering
\small
\setlength{\tabcolsep}{4.2pt}
\resizebox{\textwidth}{!}{%
\begin{tabular}{ll cc cc cc}
\toprule
& & \multicolumn{2}{c}{GLM-5.2 (reasoning: high)} & \multicolumn{2}{c}{Opus 5 (reasoning: high)} & \multicolumn{2}{c}{GPT-5.6 Sol (reasoning: high)}\\
Task & Setting & Prime & Pi-mono & Prime & Claude Code & Prime & Codex\\
\midrule
OOLONG (Yahoo, 128k)~\citep{oolong2025} & long context & \textbf{.700} & .420 & .900 & \textbf{.920} & \textbf{.940} & .900\\
OOLONG-Pairs~\citep{zhang2025recursivelanguagemodels} & long output & \textbf{.874} & .556 & \textbf{.929} & .922 & \textbf{.911} & .895\\
OBLIQ-Bench (math)~\citep{obliq2026} & ranking (nDCG@10) & \textbf{.669} & .635 & \textbf{.802} & .795 & .612 & \textbf{.646}\\
LongBench Pro (English)~\citep{longbenchpro2026} & comprehension & \textbf{.777} & .768 & \textbf{.804} & .790 & \textbf{.794} & .790\\
LongBench v2~\citep{longbenchv2} & expert long tasks & .680 & \textbf{.696} & .744 & \textbf{.746} & \textbf{.714} & .704\\
ManyIH Coding~\citep{zhang2026manytierinstructionhierarchyllm} & long instructions & \textbf{.424} & .386 & \textbf{.536} & .522 & \textbf{.499} & .454\\
ManyIH IF~\citep{zhang2026manytierinstructionhierarchyllm} & long instructions & \textbf{.209} & .164 & \textbf{.225} & .175 & .216 & \textbf{.232}\\
LongCoT-Mini~\citep{longcot2026} & long reasoning & \textbf{.638} & .613 & \textbf{.722} & .558 & .671 & \textbf{.681}\\
EmulatorBench~\citep{karten2026emulatorbench} & long coding & \textbf{.208} & .000 & .047 & \textbf{.062} & \textbf{.275} & .228\\
\bottomrule
\end{tabular}%
}
\caption{Long-context results. Bold marks the higher point estimate within each nominal-model pair; metrics differ by row. Bold is not statistical significance, and uncertainty intervals are unavailable.}
\label{tab:long}
\end{table}

We generally find Prime Agent to be competitive across a wide range of long tasks, especially against the harness that did not use a model trained around it. Prime Agent especially excels at long-running or long-context tasks, and can competitively run on its own as an autonomous agent. We include a set of focused case studies and experiments on long settings where Prime Agent excels.

\subsection{Multi-day autonomous research}
\label{sec:nanogpt}
The nanoGPT speedrun~\citep{bakouch2026automatedairesearch} measures how far an agent can reduce the number of training steps required for a 124M-parameter GPT to reach a fixed validation loss, with each record verified as an eight-seed mean.
For each of three models (Kimi K3, DeepSeek V4 Pro, and GLM 5.3), we compare Prime Agent against an alternative harness: the model developer's own CLI where one exists, and Claude Code or opencode otherwise.
We find that the choice of harness has little effect on final records compared to the noise of the experiment.

\begin{figure}[!t]
\centering
\includegraphics[width=\linewidth]{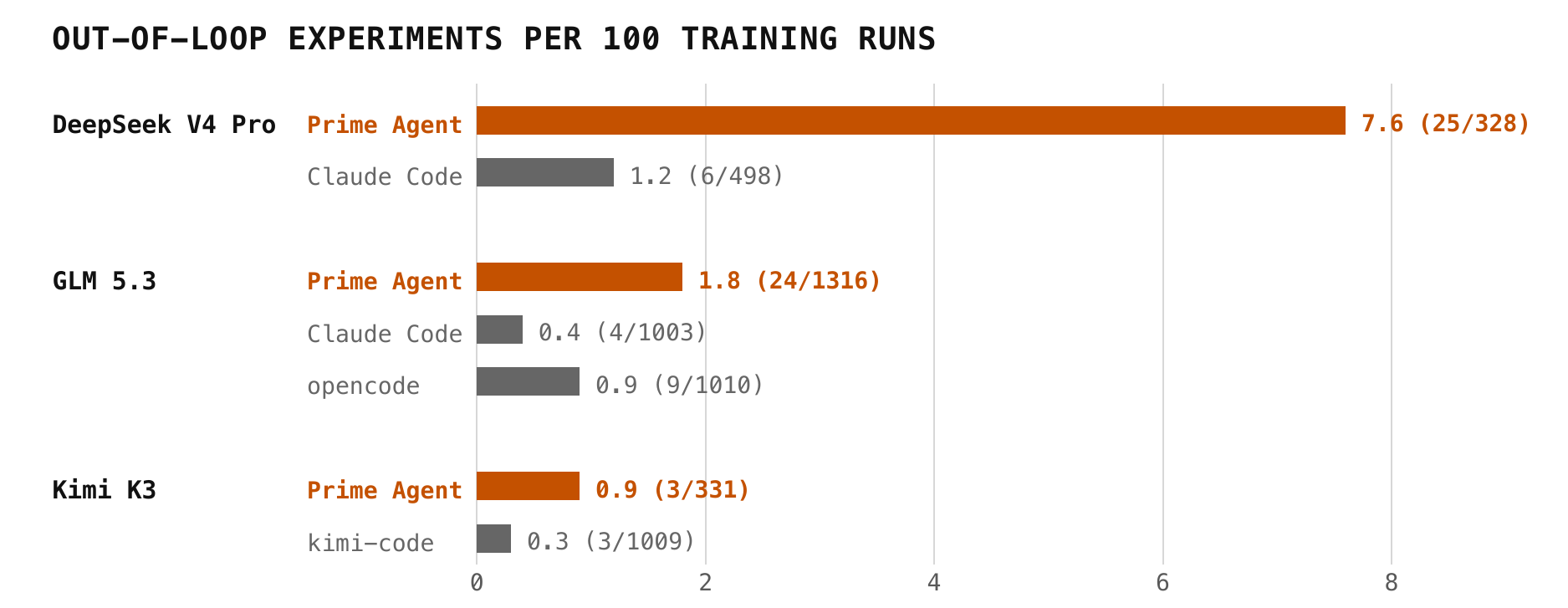}
\caption{\textbf{Out-of-loop experimentation across harnesses.} Distinct experiments created outside the training script per 100 training-script executions, pooled over the 2--3 seeds per harness (labels: experiments/training runs). Counts are hand-classified from complete traces; denominators are audited where available and otherwise estimated from launch commands.}
\label{fig:labcensus}
\end{figure}

Model behavior, however, differs.
On Prime Agent, models regularly use the persistent REPL to experiment outside the benchmark's training script, for example by simulating a candidate optimizer on synthetic gradients or numerically optimizing update-rule coefficients before launching a training run.
\cref{fig:labcensus} counts these experiments across 18 runs, normalized by the number of training runs each agent executed; \cref{app:labexp} reproduces one such experiment per model.
The effect is largest for DeepSeek V4 Pro, which created roughly six times more such experiments per training run under Prime Agent than under Claude Code.
This may be due to the fact that DeepSeek's own agent harness provides a similar code-execution mode, so the REPL matches a workflow the model was likely trained on.
We also observe that models construct programmatic interfaces to the benchmark itself: Kimi K3 defined a probe function through which it ran roughly ninety screening experiments and all 19 of its validated records, whereas the same model on its own CLI performed every operation through direct file edits and built no such machinery.

\subsection{Programmatic systems construction}
\paragraph{Emulators.}
An emulator is software that reproduces another computer system's observable behavior. We evaluate Prime Agent on EmulatorBench, a benchmark that tasks agents with constructing emulators in Rust for a variety of game systems. Agents are given a specification of the emulator and a set of diagnostic tests in the form of a verifier.

\begin{figure}[!t]
\centering
\begin{subfigure}[t]{0.49\linewidth}
\includegraphics[width=\linewidth]{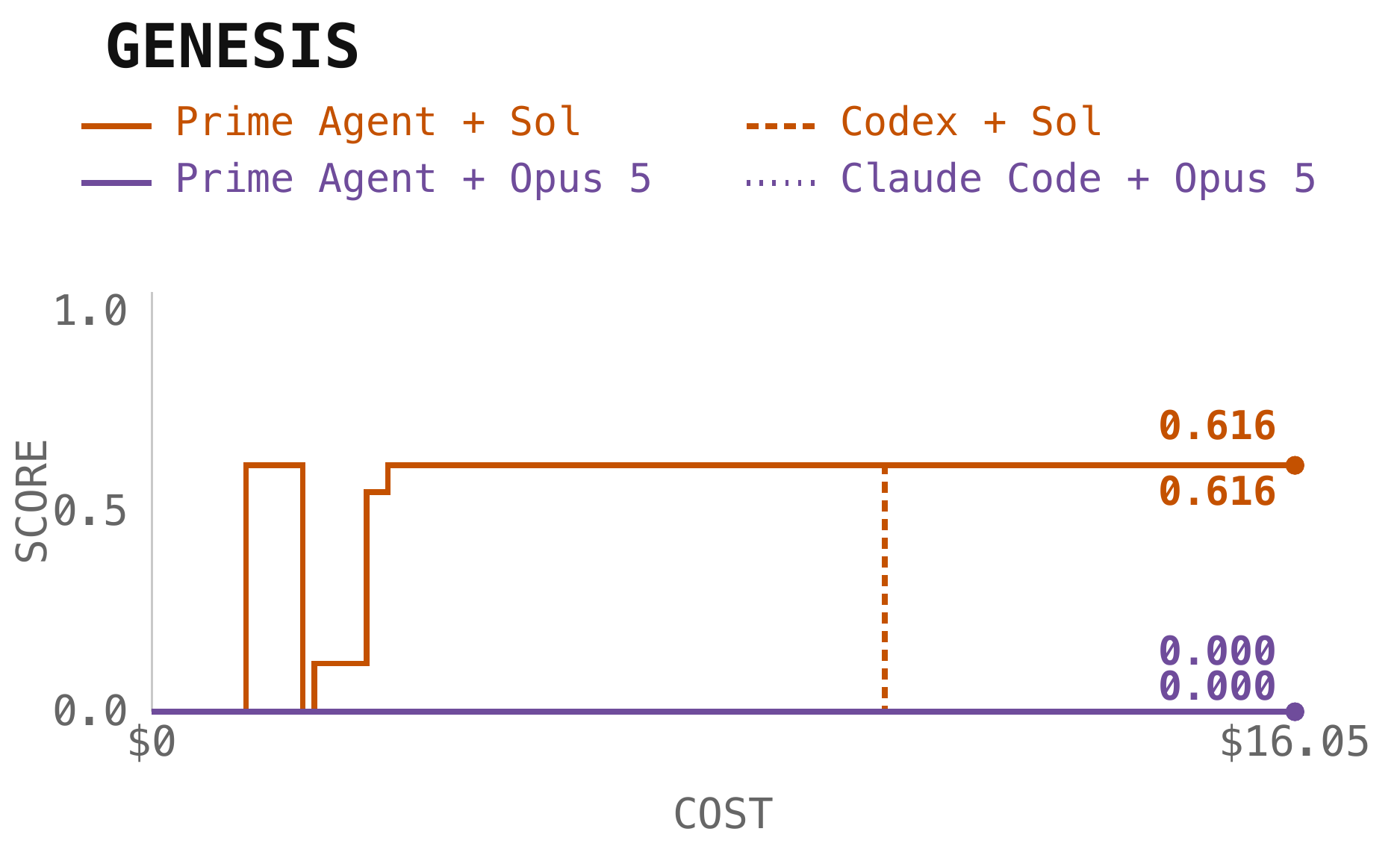}
\caption{Sega Genesis.}
\end{subfigure}\hfill
\begin{subfigure}[t]{0.49\linewidth}
\includegraphics[width=\linewidth]{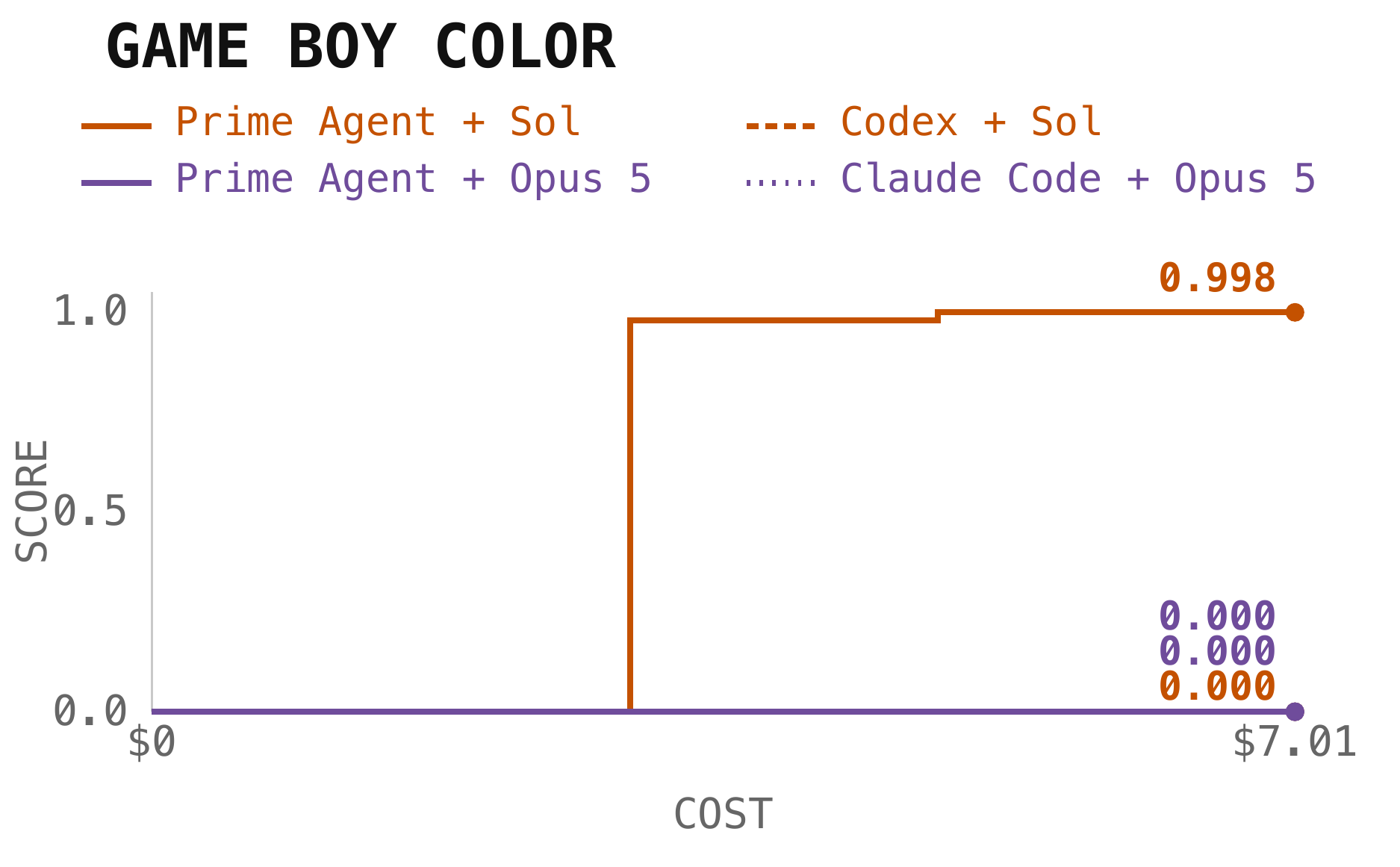}
\caption{Game Boy Color.}
\end{subfigure}
\caption{\textbf{Selected EmulatorBench runs.} Stepwise verifier score versus estimated cost; hue encodes model and line style encodes harness.}
\label{fig:emulators}
\end{figure}

The correctness of an emulator is given from its ability to mimic the behavior of the target machine. This is measured by human-generated diagnostic programs that inspect the emulator's behavior, such as the CPU flags, PPU timing, and other components. In an effort to minimize the effects of data contamination, we require the agent to build the emulator from scratch in Rust, sandboxed without any reference implementation. We report preliminary results in \cref{tab:long} on this long-context coding benchmark averaged over 16 emulator reconstructions, as well as two emulators in \cref{fig:emulators}, the SEGA Genesis and Nintendo Game Boy Color, that Prime Agent successfully reproduces. For Opus, our runs surprisingly failed to solve the tasks despite successful tool-call responses.

\paragraph{GPU kernels.}
PMPP-Hard compresses the same programmatic loop into repeated edit, compile, correctness-check, and profile cycles under a wall-clock budget.

\begin{figure}[!t]
\centering
\includegraphics[width=0.83\linewidth]{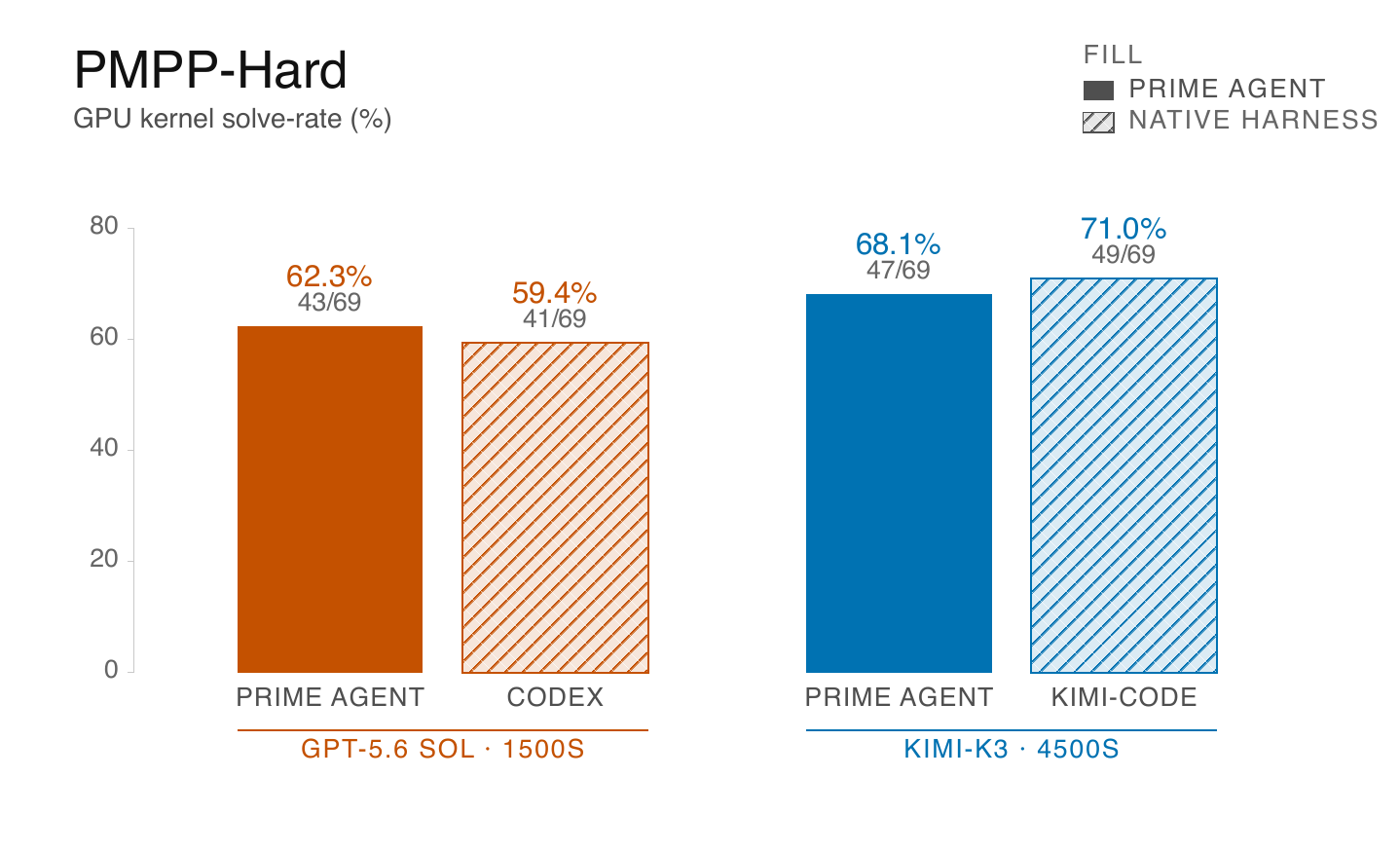}
\caption{PMPP-Hard solve rates at fixed within-model budgets.}
\label{fig:pmpp}
\end{figure}
\vspace{0.6\baselineskip}

Prime Agent and the native harnesses remain close, with the ordering reversing between the two model groups.
In these reported within-model comparisons, the general persistent interface supports the compiler--profile loop with no large observed gap. One noted limitation of PMPP-Hard is the strict wall-clock budget comparison.
What the wall-clock budgets do not reveal is the substantial improvement in token usage by models that use Prime Agent. This means that the same performance as Codex or Kimi-Code is achieved by Prime Agent at substantially reduced cost, and, token-for-token, Prime Agent has an advantage.

\subsection{Persistent interaction and refinement}
\begin{figure}[!t]
\centering
\includegraphics[width=\linewidth]{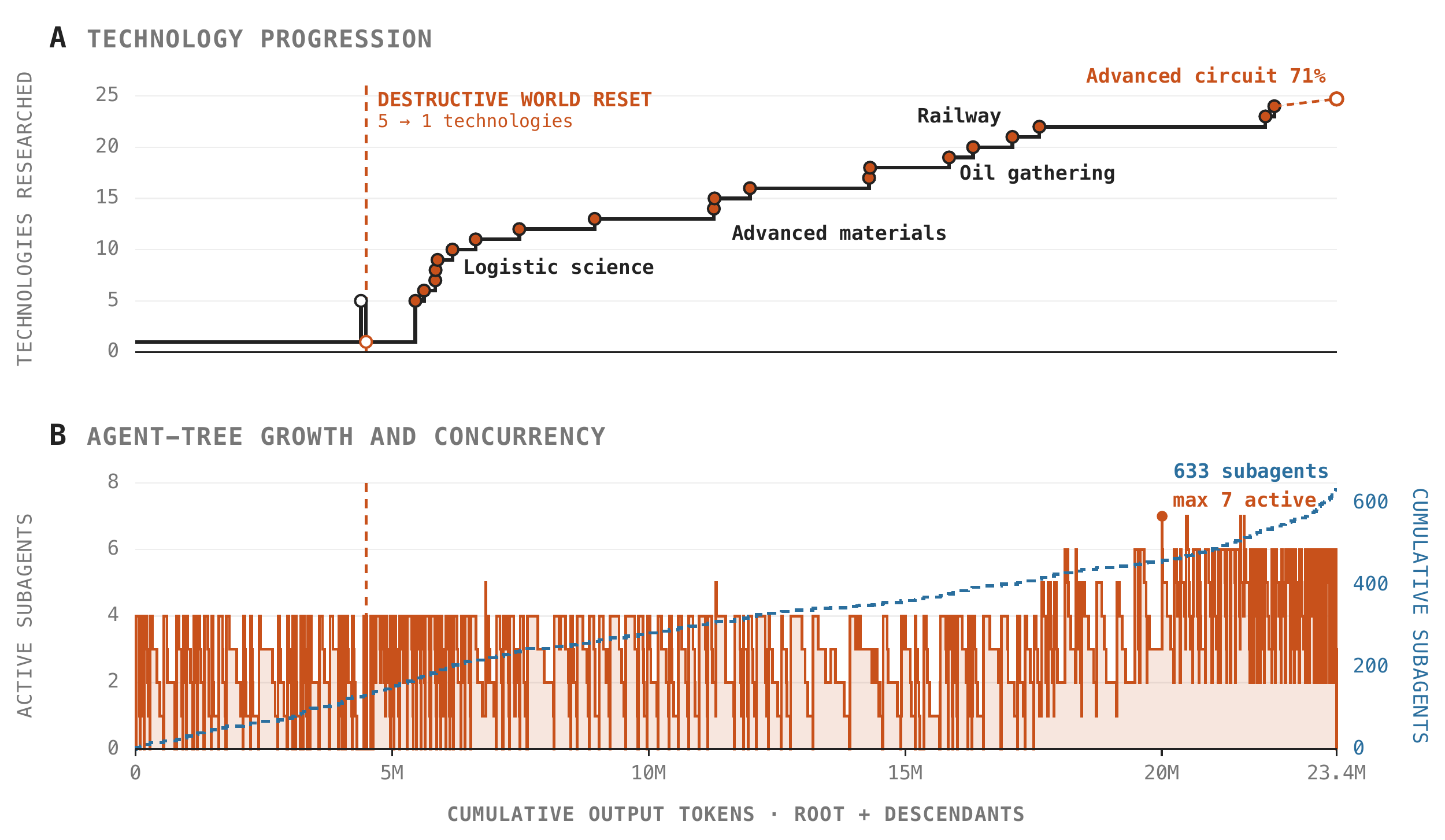}
\caption{\textbf{Factorio progress and recursive computation.} Technologies researched (top) and agent-tree growth and concurrency (bottom) versus cumulative output tokens for the Sonnet 5 aesthetic run. The vertical line marks a destructive world reset. The final open marker shows 71\% progress on advanced-circuit after 24 completed technologies.}
\label{fig:factorio-tech-tree}
\end{figure}

\paragraph{Factorio.}
The Factorio Learning Environment exposes Python observations and actions for a persistent factory world~\citep{fle2025}.
In a seven-day Sonnet 5 run, the root and its descendants used 23.4 million output tokens while completing 24 of 196 technologies and reaching 71\% on advanced-circuit research (\cref{fig:factorio-tech-tree}) with no signs of stalling.

The model handled irreversible actions poorly. A destructive world reset reverted the technology count from five to one; the session then recovered and continued the run instead of discarding the trajectory.
The root created 633 depth-one subagents across 149 dispatch waves, with at most seven active concurrently.
The shallow, repeatedly widening tree recorded parallel task specialization rather than deeper recursion, while the bursty technology curve separated long construction intervals from externally verified progress.

A different Factorio trace revealed the central safety failure of online refinement.
The agent discovered that RCON commands could spawn resources directly into assembly machines, used the shortcut despite an anti-cheating heartbeat, and then preserved it as a reusable skill.
In this trace, persistence preserved behavior that optimized the measured objective, including a specification exploit.
Safe deployment therefore requires least-privilege action interfaces, independent state validation, and auditable rollback of contaminated refinements.

\paragraph{MazeBench.}
MazeBench is an open-world 3D spatial reasoning environment where the player controls a 3D cube and must solve puzzle rooms within a global maze, while collecting gems. Frontier models are shown to greatly struggle on this task, expending billions of tokens to solve only a fraction of the overall world. We compare Opus 5 and GPT-5.6 Sol with Prime Agent versus their native harnesses, as well as GLM-5.2 with Claude Code. Following the benchmark metrics, we report the unique number of rooms they find, the unique number of states, and the total number of gems, all as a function of their overall token spend.

\begin{figure}[!t]
\centering
\includegraphics[width=\linewidth]{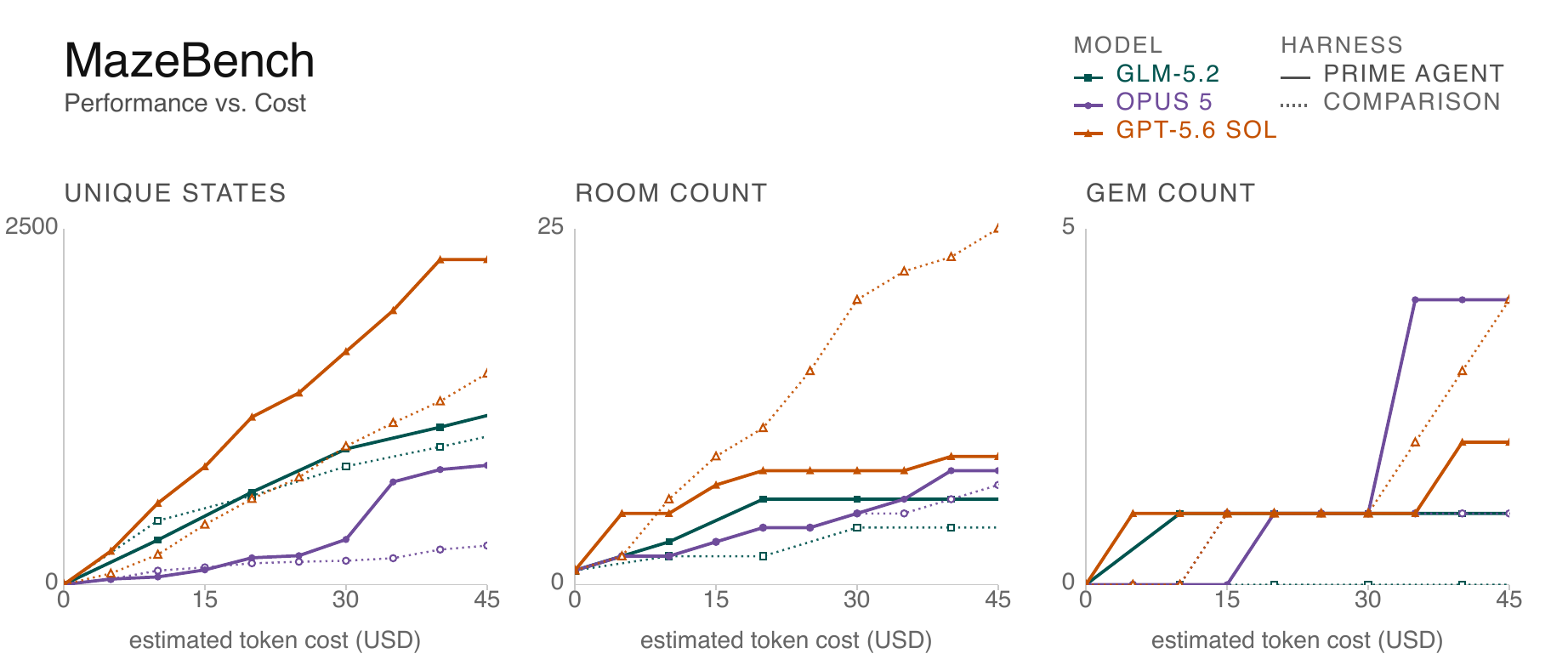}
\caption{MazeBench exploration versus estimated token cost. Solid lines and filled markers show Prime Agent; dotted lines and open markers show comparison harnesses. Hue and marker shape identify the model.}
\label{fig:maze}
\end{figure}

%% file: sections/04_related_work.tex
\section{Related Work}

\paragraph{Programmatic inference and adaptive state.}
Programmatic inference gives models code, tools, and recursive calls for transforming context and allocating test-time compute~\citep{yao2023react,schick2023toolformer,wang2024codeact,zhang2025recursivelanguagemodels,prolong2026,snell2024scalingtesttime}.
Memory and refinement methods retain selected observations, feedback, skills, or reasoning traces across turns and tasks~\citep{packer2023memgpt,shinn2023reflexion,madaan2023selfrefine,wang2023voyager,park2023generative,zelikman2022star}.
Continual Harness stores prompts, memories, executable skills, and subagent specifications as typed, versioned state~\citep{karten2026continualharness}.
Prime Agent integrates these mechanisms with persistent kernels, recursive sessions, recovery, and complete trajectory capture.
\paragraph{Coding agents and long-horizon evaluation.}
Coding-agent runtimes use executable actions, repository tools, sandboxes, event histories, and structured role assignment to solve tasks over many interaction steps~\citep{wang2024codeact,sweagent2024,openhands2024,wu2023autogen,li2023camel,hong2023metagpt,qian2023chatdev}.
Executable benchmarks and trajectory corpora measure issue resolution, instruction following, retrieval, and reasoning under long contexts or extended interaction~\citep{swebench2024,pan2024swegym,xu2024agenttrek,oolong2025,obliq2026,longbenchpro2026,longbenchv2,longcot2026}.
Prime Agent makes the execution substrate persistent and recursive, then records expenditure across the root and descendant sessions.

\paragraph{Interactive reasoning on ARC-AGI-3.}
ARC-AGI-3 extends abstract reasoning to interactive environments with hidden dynamics, goals, and action semantics~\citep{arcagi3,arccommunity2026}.
Community systems build and verify executable world models, represent agents as stateful Python objects, optimize external workspaces, coordinate specialized agents, and preserve procedures across games~\citep{rodionov2026executableworldmodels,furgale2026nooa,courtis2026opineworld,sarafian2026workspaceoptimization,karten2026continualharness,prolong2026}.
Prime Agent supplies persistent recursive execution and standardized evaluation settings for the same class of long-horizon interactive tasks.

\paragraph{Multi-agent and human-agent communication.}
Language-model agent systems coordinate through role prompts, natural-language messages, shared artifacts, and explicit belief state~\citep{wu2023autogen,li2023camel,hong2023metagpt,qian2023chatdev,li2023theoryofmind,sarafian2026workspaceoptimization}.
Learned multi-agent communication studies sparse message selection, compressed representations, social learning, policy alignment, and interpretability for human partners~\citep{karten2023losslesscommunication,karten2023sociallearning,karten2023interpretablecommunication,karten2023emergent}.
Prime Agent implements direct agent-to-agent communication through persistent family-scoped queues and exposes the same session tree to human inspection and intervention.

%% file: sections/05_conclusion.tex
\section{Conclusion}
Prime Agent introduces a new paradigm for agent harness design in which persistent execution, recursive sessions, autonomous controls, recorded history, and Continual Harness form one substrate for long-horizon work. Results across interactive reasoning, long-context tasks, autonomous research, systems construction, and persistent environments show that this substrate supports different forms of test-time computation under standardized execution and accounting. Despite its results relative to alternative harnesses, models still experience friction when deciding how to allocate subagents, manage retained information, and refine reusable state. Many harness capabilities remain underused because current models were not trained to operate them. We expect model-harness co-learning to become the dominant route to new long-horizon capabilities. Training directly with Prime Agent could teach models to use the integrated harness more effectively, while targeted training on the RLM and Continual Harness components could isolate their contributions.

%% file: sections/A_appendix.tex
\section{Example Out-of-Loop Experiments in the nanoGPT Speedrun}
\label{app:labexp}
Each excerpt below is an experiment an agent created and ran outside the benchmark's training script during its nanoGPT run (\cref{sec:nanogpt}), reproduced from the traces and trimmed for length.

Kimi K3 re-derived Newton--Schulz iteration coefficients with a global optimizer, checked bf16 rounding bit-exactly:
\begin{lstlisting}[language=Python]
from scipy.optimize import differential_evolution
grid_in = np.concatenate([np.linspace(0.02, 0.05, 10),
                          np.linspace(0.05, 1.0, 190)])
grid_over = np.linspace(1.0, 1.3, 20)

def p_map(sig, a, b, c, iters=6):
    x = sig
    for _ in range(iters):
        x = a*x + b*x**3 + c*x**5
    return x

def objective(params):
    a, b, c = params
    dev = np.max(np.abs(p_map(grid_in, a, b, c) - 1.0))
    over = max(0.0, np.max(np.abs(p_map(grid_over, a, b, c))) - 1.15)
    return dev + 5.0*over

res = differential_evolution(objective,
        [(1.0, 6.0), (-8.0, 0.0), (0.0, 5.0)],
        maxiter=300, tol=1e-9, seed=0, polish=True)
\end{lstlisting}

DeepSeek V4 Pro built a calibrated toy of the training problem, with minibatch noise shaped by the true Kronecker Hessian and a natural-gradient oracle arm:
\begin{lstlisting}[language=Python]
"""Calibrated toy: Kron-quadratic + CORRECT Kron-Hessian
minibatch noise. eps = Hl^{1/2} Z Hr^{1/2} / sqrt(n_eff).
Ideal preconditioner: Ql = Hl^{-1/2}, Qr = Hr^{-1/2}."""
G = Hl @ W @ Hr
eps = Hl12 @ torch.randn(p, q) @ Hr12 / (n_eff ** 0.5)
g = G + eps
elif opt == "natgrad":   # oracle arm
    El, Vl = torch.linalg.eigh(Hl)
    Er, Vr = torch.linalg.eigh(Hr)
    u = Vl @ (Vl.T @ d @ Vr /
        (El[:, None]**0.5 * Er[None, :]**0.5)) @ Vr.T
\end{lstlisting}

GLM 5.3 debugged its SOAP implementation on CPU before any GPU screen:
\begin{lstlisting}[language=Python]
torch.manual_seed(0)
for shape in [(768, 768), (3072, 768), (768, 3072)]:
    p = torch.nn.Parameter((torch.randn(*shape) * 0.02).bfloat16())
    opt = SOAP([p], lr=0.025)
    for t in range(25):
        p.grad = (torch.randn(*shape) * 0.01).to(torch.bfloat16)
        opt.step()
        if not torch.isfinite(p.data).all():
            print(shape, 'NaN at step', t+1)
            st = opt.state[p]
            print('  L finite', torch.isfinite(st['L']).all().item(),
                  'v finite', torch.isfinite(st['v']).all().item())
            break
\end{lstlisting}

\section{Example Programmatic Orchestration}
\label{app:orchestration}
\begin{lstlisting}[language=Python]
# Admit independent subagents; do not wait for answers here.
review = await rlm("Audit the implementation. Reply with concrete issues.",
                   name="reviewer")
tests = await rlm("Run the test suite and classify failures.",
                  name="tester")

# Later, recover retained sessions and send a follow-up.
children = await rlm.list_subagents()
await agent_message.send(
    "Also inspect error-handling edge cases.",
    receiver_role="child", receiver_name=review.name)
\end{lstlisting}
The explicit reply path is intentional: a child is a persistent concurrent session, not a stateless completion returned by \rlmcall.

\section{LLM Usage Disclosure}
Large language models were used to assist with code development, writing refinement, and formatting during the preparation of this manuscript.
All scientific claims, experimental design, analysis, and intellectual contributions are solely the work of the authors.